\documentclass[10pt,twocolumn,letterpaper]{article}

\usepackage[pagenumbers]{cvpr} 
\usepackage{graphicx} 
\usepackage{adjustbox} 
\definecolor{cvprblue}{rgb}{0.21,0.49,0.74}
\usepackage[pagebackref,breaklinks,colorlinks,allcolors=cvprblue]{hyperref}
\usepackage[normalem]{ulem}

\title{Text to Trust: Evaluating Fine-Tuning and LoRA Trade-offs in Language Models for Unfair Terms of Service Detection}

\author{
Noshitha Padma Pratyusha Juttu\textsuperscript{1} ,
Sahithi Singireddy\textsuperscript{1} ,
Sravani Gona\textsuperscript{1},
Sujal Timilsina\textsuperscript{1}
\\
\textsuperscript{1}\textit{University of Massachusetts Amherst}
}

\begin{document}
\maketitle
\begin{abstract}
Terms of Service (ToS) agreements often contain clauses that are difficult to interpret and potentially unfair to users. Manual identification of such clauses is infeasible at scale, motivating the need for automated, accurate, and efficient detection methods. This study presents a comprehensive evaluation of clause-level unfairness detection using a diverse range of large language model (LLM) strategies, including full fine-tuning, parameter-efficient tuning, and zero-shot prompting. Experiments are conducted with full fine-tuning on BERT and DistilBERT, 4-bit quantized Low-Rank Adaptation (LoRA) applied to models such as TinyLlama and LLaMA, and to the legal domain-specific SaulLM, and evaluate zero-shot prompting using high-performing API-accessible models like GPT-4o and O3-mini. Evaluations are performed on the Claudette-ToS dataset from Hugging Face and further validated on the Multilingual Scraper of Privacy Policies and Terms of Service corpus, which comprises large-scale ToS documents collected from the web. Full fine-tuning delivers the strongest overall performance, parameter-efficient models offer a favorable accuracy–efficiency trade-off, and zero-shot prompting enables fast deployment with high recall. These results offer practical insights into building scalable and cost-effective unfairness detection systems for legal-tech applications.
\end{abstract}

\section{Introduction}
Terms of Service (ToS) agreements are ubiquitous in the digital age, governing nearly every interaction between users and online platforms. Despite their legal importance, these documents are typically long, densely worded, and difficult for lay users to interpret. As a result, users routinely accept terms without reading or understanding them, unknowingly agreeing to clauses that may include liability waivers, forced arbitration, unilateral modifications, or the loss of important legal rights. Identifying such unfair or predatory clauses is critical not only for empowering consumers but also for enabling regulatory agencies and watchdog groups to audit platform behavior at scale.

Traditional legal review is manual, expensive, and not feasible at the volume and velocity required by today’s internet-scale services. Automated clause-level classification using language models has therefore emerged as a promising solution. Prior work has shown that transformer-based architectures, especially models like BERT fine-tuned on labeled legal datasets, can perform well in detecting unfairness. However, full fine-tuning of large models is computationally intensive, limits scalability, and requires substantial labeled data. In contrast, smaller or rule-based models often lack the semantic depth to capture nuanced unfairness, particularly when phrased in indirect or ambiguous legal language.

This study presents a unified and comprehensive evaluation of automated unfairness detection in ToS clauses using three complementary modeling paradigms: 
\begin{enumerate}
    \item Full fine-tuning of pretrained transformers such as BERT and DistilBERT,
    \item Parameter-efficient fine-tuning using Low-Rank Adaptation (LoRA) with 4-bit quantization applied to small and mid-sized models like TinyLlama, LLaMA, and SaulLM, and
    \item Zero-shot prompting using state-of-the-art API-accessible Large Language Models including GPT-4o and o3-mini.
\end{enumerate}
Each paradigm represents a distinct trade-off in accuracy, generalization, and resource efficiency.

These models are evaluated on the Claudette-ToS dataset, a benchmark comprising thousands of annotated clauses labeled as fair or unfair. To assess real-world applicability, the best-performing models are further deployed on a large multilingual ToS corpus collected from thousands of websites. This large-scale evaluation tests generalization beyond curated benchmarks to naturally occurring, diverse legal language.

By conducting a detailed comparison across fine-tuned, parameter-efficient, and zero-shot approaches, this work provides practical insight into the design of scalable unfair clause detectors. It contributes to the growing intersection of legal informatics and NLP by demonstrating how modern language models can support automated compliance monitoring, user transparency, and large-scale legal document auditing.

\section{Related Work}

The problem of detecting unfair clauses in Terms of Service (ToS) agreements has long posed challenges to both the legal and NLP communities. Prior research has explored feature-driven classification, transformer-based models, domain-specific LLMs, and scalable fine-tuning. This work builds upon these threads by providing a comparative evaluation across full fine-tuning, parameter-efficient LoRA tuning, and zero-shot prompting, with an emphasis on real-world, web-scale ToS data.

\textbf{Early approaches using feature-based methods:}
Lippi et al.~\citep{lippi2019claudette}  introduced CLAUDETTE, a pioneering system for detecting potentially unfair clauses. They manually annotated 50 ToS agreements with multiple clause categories and trained traditional classifiers (SVMs, CNNs, LSTMs), finding that shallow lexical features often outperformed more complex syntactic structures.  Their annotated dataset was later expanded into a larger binary-labeled benchmark, released on Hugging Face as \texttt{claudette\_tos}~\citep{claudette2024dataset}, which serves as a primary benchmark for fairness classification.

\textbf{Transformer-based advances and domain adaptation:}
BERT~\citep{devlin2019bert} revolutionized clause classification by enabling deep contextual understanding of legal language. Chalkidis et al.~\citep{chalkidis2020legalbert} adapted BERT for the legal domain with Legal-BERT, while Bathini et al.~\citep{bathini2023unfair} fine-tuned Legal-BERT on UNFAIR-ToS, achieving an F1 score above 0.92.

\textbf{Scaling legal understanding via domain-specific LLMs:}
As transformer models have scaled, domain-specialized LLMs have emerged. Colombo et al.~\citep{colombo2024saullm} introduced SaulLM, a 7B-parameter model pre-trained on 19 million legal documents, showing strong results on legal QA and clause classification tasks. These models highlight trade-offs between domain specificity, resource requirements, and fine-tuning efficiency. 

\textbf{Parameter-efficient fine-tuning:}
LoRA~\citep{hu2022lora} introduced low-rank matrices in frozen transformers, while QLoRA~\citep{dettmers2023qlora} combined it with 4-bit quantization, reducing memory usage with minimal accuracy loss. Such techniques have been extended to models like SaulLM-7B, LLaMA-3B/7B, and TinyLlama-1.1B, offering insights into where LoRA-tuned models excel (e.g., recall) and where they fall short (e.g., nuanced fairness detection).

\textbf{Zero-shot prompting and model reliability:}
Recent studies have examined zero-shot inference for legal clause detection. Zhou et al.~\citep{zhou2024llmfailures} showed that general-purpose LLMs often misclassify due to limited domain grounding. While GPT-4 variants achieve high recall, they typically exhibit lower precision in legal tasks, underscoring the need for targeted evaluation frameworks.

\textbf{Real-world evaluation on large-scale web corpora:}
Most prior work focuses on curated benchmarks. Bernhard et al.~\citep{bernhard2024multilingualtos} addressed this limitation by constructing a multilingual corpus of over 20,000 ToS and privacy policies. This corpus enables large-scale deployment and analysis of unfairness detection systems in diverse, naturally occurring legal text.

\textbf{Positioning of this work:}
Unlike prior studies that focus on a single modeling paradigm (e.g., BERT fine-tuning or GPT prompting), this paper presents a unified comparison across model scale, fine-tuning strategy, and inference paradigm. It further extends existing research by deploying fairness classifiers on web-scale ToS data, offering insights into model robustness and trade-offs relevant to both NLP research and legal-tech applications.

\section{Datasets}

To develop and evaluate models for detecting unfair clauses in Terms of Service (ToS) documents, two complementary datasets are utilized.The first is a curated benchmark with clause-level annotations, ideal for supervised fine-tuning and evaluation. The second is a large, real-world scraped corpus designed to assess model generalization and robustness in practical deployment settings.

\subsection{CLAUDETTE-ToS Dataset}

The primary dataset for model training and supervised evaluation is the \textbf{CLAUDETTE-ToS dataset}~\citep{claudette2024dataset}, which contains 9,414 English clauses extracted from real-world online contracts. Each clause is manually labeled as either \textit{fair} or \textit{unfair}, enabling precise clause-level classification.

The original distribution is skewed, with 8382 clauses (89.1\%) labeled as fair and 1032 (10.9\%) labeled as unfair. To mitigate this imbalance, a balanced subset was constructed by randomly sampling an equal number of clauses from each class. The preprocessing pipeline included tokenization using the target model’s tokenizer and a maximum clause length of 256 tokens. The dataset was split into 80\% training, 10\% validation, and 10\% testing subsets.

Example clauses include:
\begin{itemize}
\item \textbf{Fair (Label 0):} "Please check the latest rates before you make your call."
\item \textbf{Unfair (Label 1):} "The company may change the rates for calling phones at any time without notice to you by posting such change in the website." 
\end{itemize}

The dataset\footnote{CLAUDETTE-ToS Dataset:\url{https://huggingface.co/datasets/LawInformedAI/claudette_tos}} is publicly available on Hugging Face under an open license and has been widely adopted for ToS fairness classification tasks.

\subsection{Multilingual Scraped ToS Corpus}
To evaluate the generalization of the model in practical scenarios,the \textbf{Multilingual Scraper of Privacy Policies and Terms of Service}~\citep{bernhard2024multilingualtos} dataset was used. This large-scale dataset containing approximately 60GB of HTML content scraped from thousands of websites over 12 months.

Each scraped document is accompanied by rich metadata that aids in filtering and analysis. Key fields include:

\begin{itemize}
  \item \textbf{Document identifiers:} \texttt{term\_url\_id}, \texttt{website\_month\_id}
  \item \textbf{Language indicators:} \texttt{website\_lang}, \texttt{term\_lang}
  \item \textbf{Scoring attributes:} \texttt{term\_keyword\_score} (keyword-based heuristic), \texttt{term\_ml\_probability} (model confidence that the text is ToS)
  \item \textbf{Content fields:} \texttt{term\_url}, \texttt{term\_content}, \texttt{term\_content\_hash}
\end{itemize}

For evaluation, only English-language documents were retained (\texttt{website\_lang = term\_lang = en}). The remaining metadata was used for filtering and analysis, focusing on documents likely to be authentic ToS pages. This corpus provided the basis for large-scale deployment testing and assessment of unfairness detection models under realistic, noisy web conditions.

\subsection{Ethical and Licensing Considerations}
Both datasets are publicly released for research purposes and do not contain personal or sensitive user data. CLAUDETTE-ToS is distributed under a permissive license via Hugging Face, and the Multilingual Scraper corpus is accessible through the ACM Digital Library. All dataset usage in this work complies with their respective licensing terms.

\section{Methodology}

This section outlines the modeling paradigms, architectures, and implementation pipelines used to evaluate clause-level unfairness detection. Three complementary approaches were investigated: 
(1) full fine-tuning of encoder-based transformers, 
(2) parameter-efficient adaptation through Low-Rank Adaptation (LoRA) with quantization, and
(3) zero-shot prompting using instruction-tuned large language models.

\subsection{Model Architectures}

\subsubsection{Fully Fine-Tuned Transformers}
BERT~\citep{devlin2019bert} and DistilBERT serve as encoder-based baselines for full fine-tuning. BERT (110M parameters) provides strong contextual representations, while DistilBERT offers comparable performance with roughly 40\% fewer parameters. Both models were fine-tuned end-to-end for binary classification of clauses labeled as \textit{fair} or \textit{unfair}. These models establish performance ceilings for smaller or more efficient variants.

\subsubsection{Parameter-Efficient Fine-Tuning (PEFT)}
Parameter-efficient adaptation was implemented using Low-Rank Adaptation (LoRA)~\citep{hu2022lora} combined with 4-bit quantization following QLoRA~\citep{dettmers2023qlora}.  
Four models were examined:
\begin{itemize}
  \item \textbf{TinyLlama-1.1B:} A compact model optimized for chat-style tasks. LoRA adapters were injected into the query and value projections under 4-bit quantization (NF4). This configuration explores the feasibility of ultra-lightweight models for legal-text classification.
  \item \textbf{LLaMA-3B and LLaMA-7B:} Mid-sized open models evaluated under quantized LoRA adaptation. Due to GPU resource constraints, training was limited to one or two epochs but provides insight into scaling trends.
  \item \textbf{SaulLM-7B:} A legal-domain-specific model pre-trained on over 19 million legal documents~\citep{colombo2024saullm}. LoRA adapters (rank = 16) were trained under NF4 quantization with gradient checkpointing to manage memory. Its domain specialization enables comparison between general-purpose and legally grounded LLMs.\footnote{Fine-tuned models TinyLlama-ToS and SaulLM-ToS are publicly available at \url{https://huggingface.co/Noshitha98}.}

\end{itemize}

\begin{table}[hpp]
\centering
\label{tab:hyperparams_finetune}
\resizebox{\columnwidth}{!}{%
\begin{tabular}{|l|c|c|c|c|c|c|}
\hline
\textbf{Model} & \textbf{Params} & \textbf{Tuning Method} & \textbf{Epochs} & \textbf{Batch Size} & \textbf{LR} & \textbf{Token Limit} \\
\hline
BERT           & 110M   & Full FT       & 3      & 8          & 2e-5           & 512         \\
DistilBERT     & 66M    & Full FT       & 3      & 8          & 2e-5           & 512         \\
TinyLlama-1.1B & 1.1B   & LoRA (NF4)    & 3      & 2$\times$4 & 2e-4           & 256         \\
LLaMA-3B       & 3B     & LoRA (FP4)    & 2      & 2$\times$4 & 2e-4           & 256         \\
LLaMA-7B       & 7B     & LoRA (FP4)    & 1      & 2$\times$4 & 2e-4           & 256         \\
SaulLM-7B      & 7B     & LoRA (NF4)    & 1      & 2$\times$4 & 5e-5           & 256         \\
\hline
\end{tabular}%
}
\caption{Training Hyperparameters for Fine-Tuning Baselines}
\end{table}

\subsection{Training and Implementation Pipeline}

All experiments were implemented in \texttt{Python} using the Hugging Face \texttt{transformers}, \texttt{datasets}, and \texttt{peft} libraries along with \texttt{bitsandbytes} for quantization.  
Fine-tuning was conducted on NVIDIA A100 and V100 GPUs using the SLURM cluster. Gradient accumulation and checkpointing were used to accommodate 7B-parameter models within 24 GB of VRAM.

\subsubsection{Optimization and Stability}
For LoRA models, adapter rank and learning rate were tuned to balance convergence speed and stability under quantization. Full fine-tuned transformers (BERT/DistilBERT) used cross-entropy loss with AdamW optimization and early stopping based on validation F1 score.

\subsection{Zero-Shot Prompting}

To complement supervised fine-tuning, zero-shot prompting was evaluated using proprietary instruction-tuned models accessible via the OpenAI API. Models included GPT-4o, GPT-4o-mini, O1-mini, O3-mini, and O4-mini.  
Each model received batches of five clauses formatted with a standardized instruction-style prompt:

\begin{figure}[ht]
\centering
\includegraphics[width=0.85\linewidth]{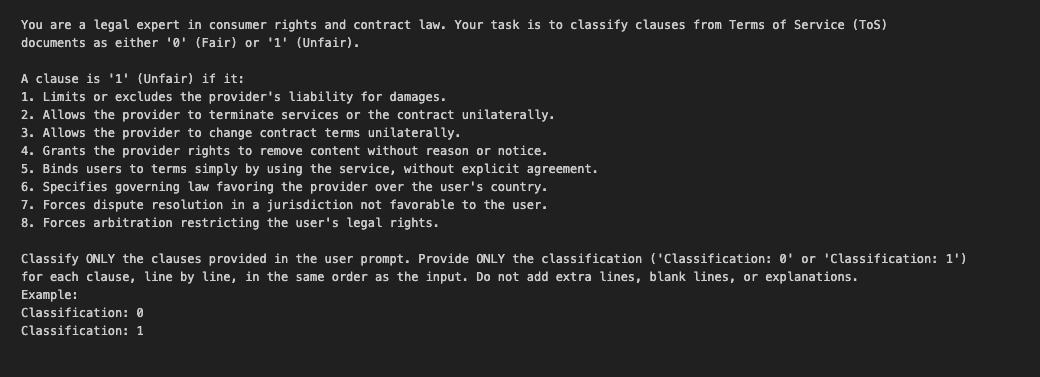}
\caption{System Prompt for zero-shot prompting}
\label{fig:zero_shot_prompt}
\end{figure}

Model responses were post-processed to binary labels and evaluated on the same held-out Claudette-ToS test set used for fine-tuned models. This setup allows comparison of resource-free inference against trained baselines in terms of precision, recall, and F1.

\subsection{Summary of Methodology}

The overall methodology integrates three distinct approaches, full fine-tuning, parameter-efficient adaptation, and zero-shot prompting, to capture trade-offs in accuracy, scalability, and computational cost. This framework enables systematic benchmarking of LLM-based clause classification across diverse model sizes and inference paradigms.

\section{Results}

This section presents the experimental outcomes of all modeling paradigms. 
Performance is reported using accuracy, precision, recall, and F1 score on the held-out Claudette-ToS test set. 
Additional deployment-scale results are discussed in Section~\ref{sec:deployment}.

\subsection{Quantitative Evaluation of Fine-Tuned Models}

The fully fine-tuned transformers, BERT and DistilBERT set the performance ceiling, achieving near identical F1 scores of 89.2\% and 89.6\%, respectively. These models maintain an excellent precision–recall balance, demonstrating the strength of full-parameter optimization when ample labeled data and computation are available.

In contrast, the parameter-efficient LoRA-adapted models yield more nuanced trade-offs. The legal domain SaulLM-7B achieves a recall of 97.5\%, substantially higher than the fully fine-tuned baselines, but its precision drops to 73.6\%, indicating broader clause coverage at the cost of specificity. The smaller TinyLlama-1.1B exhibits the opposite trend of high precision (89.1\%) but modest recall (52.5\%) which illustrating compact, quantized adapters can preserve precision under tight resource budgets yet struggle to generalize. Mid-sized LLaMA-3B/7B variants fall below both groups, underscoring that parameter scale alone does not guarantee competitive fairness detection without domain specific adaptation.

Overall, these results suggest that parameter-efficient LoRA tuning approaches partial parity with full fine tuning, achieving competitive recall and acceptable precision despite significantly lower training cost and memory usage. However, fully fine-tuned models remain the most balanced and reliable option when accuracy and calibration are critical for downstream compliance auditing.

\begin{table}[h]
\centering
\label{tab:fine_tuned_results}
\resizebox{\columnwidth}{!}{%
\begin{tabular}{|l|c|c|c|c|}
\hline
\textbf{Model} & \textbf{Accuracy (\%)} & \textbf{Precision (\%)} & \textbf{Recall (\%)} & \textbf{F1 (\%)} \\
\hline
BERT            & 88.86 & 89.20 & 89.20 & 89.20 \\
DistilBERT      & 89.00 & 89.87 & 89.31 & 89.58 \\
TinyLlama-1.1B  & 73.02 & 89.05 & 52.50 & 66.06 \\
LLaMA-3B        & 57.82 & 58.20 & 59.64 & 58.91 \\
LLaMA-7B        & 58.03 & 59.06 & 58.05 & 58.55 \\
SaulLM-7B       & 82.25 & 73.58 & 97.50 & 83.87 \\
\hline
\end{tabular}%
}
\caption{Test-set performance of fine-tuned models on the Claudette-ToS dataset.}
\end{table}
This presents the quantitative performance of all fine-tuned baselines on the Claudette-ToS test set.

\subsection{Zero-Shot Evaluation}

To assess model performance without any task-specific training, five proprietary instruction-tuned models were evaluated in a zero-shot setting. All models exhibit high recall ($>89\%$) but relatively low precision, suggesting that they capture most unfair clauses yet over-flag borderline cases.

\begin{table}[h]
\centering
\label{tab: zero_shot_prompt_results}
\resizebox{\columnwidth}{!}{%
\begin{tabular}{|l|c|c|c|c|}
\hline
\textbf{Model} & \textbf{Accuracy} & \textbf{Precision} & \textbf{Recall} & \textbf{F1 Score} \\
\hline
GPT-4o         & 75.27\%       & 29.30\%     & 89.32\%  & 44.12\%    \\
GPT-4o-mini    & 78.03\%       & 32.07\%     & 90.29\%  & 47.33\%    \\
O1-mini        & 74.52\%       & 29.05\%     & 92.23\%  & 44.19\%    \\
O3-mini        & 85.67\%       & 42.73\%     & 91.26\%  & 58.20\%    \\
O4-mini        & 80.89\%       & 35.36\%         & 90.29\%      & 50.82\%        \\
\hline
\end{tabular}%
}
\caption{Zero-Shot Prompting Results on Claudette-ToS Test Set}
\end{table}
Among all variants, O3-mini achieves the best overall balance with an F1 score of 58.2\%, followed by GPT-4o-mini. 
The consistent high recall across models highlights the effectiveness of large instruction-tuned models in recognizing potential unfairness even without supervision.

\subsection{Classifier Selection and Analysis}

While DistilBERT marginally outperforms BERT in F1 score (89.58 \% vs.\ 89.20 \%), BERT was selected for downstream deployment (Section~\ref{sec:deployment}) due to its stable calibration and interpretability.  
BERT produced smoother probability distributions and fewer spurious activations for ambiguous clauses, making it more suitable for compliance-auditing pipelines where reliability is prioritized over minimal performance gains.

\subsection{Summary of Findings}

Overall, full fine-tuning delivers the strongest and most balanced performance.  
Parameter-efficient models such as SaulLM-7B achieve excellent recall with reduced training cost, while lightweight TinyLlama provides a viable option for edge or resource-limited inference.  
Zero-shot prompting remains attractive for rapid prototyping but lacks the precision required for production-grade legal analysis.

\section{Evaluation on Real-World Corpus}
\label{sec:deployment}

To assess model generalization beyond curated benchmarks, the best-performing fine-tuned model (BERT) was deployed on the large-scale Multilingual Scraper of Privacy Policies and Terms of Service corpus~\citep{bernhard2024multilingualtos}.  
This section examines how the classifier performs under naturally occurring, noisy web data and demonstrates its potential for large-scale legal auditing.

\subsection{Batch Inference and Data Export}

The fine-tuned BERT classifier was applied to 937 clause-level entries drawn from the English-filtered subset of the corpus.  
For each clause, the following outputs were recorded:
\begin{itemize}
    \item \texttt{predicted\_label}: 1 for \textit{unfair}, 0 for \textit{fair};
    \item \texttt{terms\_ml\_probability}: softmax confidence for the \textit{unfair} class;
    \item \texttt{terms\_keyword\_score}: crawler heuristic estimating ToS likelihood;
    \item document-level metadata (\texttt{website\_url}, \texttt{terms\_url}) for traceability.
\end{itemize}

Predictions were exported in CSV format and analyzed in \texttt{pandas} to compute distributional statistics and visualize clause-level outcomes.

\subsection{Filtering and Distribution Analysis}

Across all 937 clauses, 749 were predicted as fair and 188 as unfair.  
Model confidence scores (\texttt{terms\_ml\_probability}) ranged from 0.01 to 0.96, while keyword heuristics ranged from 0.007 to 3.51.  
To isolate likely ToS documents, clauses satisfying:

 \texttt{ml\_probability} $\ge$ 0.5 or \texttt{keyword\_score} $\ge$ 1.5.
were retained.  
This filtering produced 623 high-confidence ToS clauses.  
Of these, 152 of the 188 unfair-labeled clauses originated from high-confidence ToS sources, suggesting that roughly 80\% of the model-flagged unfair clauses correspond to authentic contractual text.

\subsection{Case Study: Example from a Public ToS}

During deployment, one publicly available Terms of Service document (from \texttt{plagramme.com}) was consistently flagged by the model as containing clauses with high unfairness probability.  
A representative clause states:

\begin{quote}
``Plagramme.com may modify the terms at any time, at its sole discretion, by updating this page. You should periodically check this page for updates. Do not use Plagramme.com if you do not agree to these Terms of Use.''
\end{quote}

This clause was \textit{predicted by the model} as potentially unfair because:
\begin{itemize}
    \item \textbf{Unilateral change without notice:} The service can update terms arbitrarily.
    \item \textbf{No user notification:} Responsibility is shifted entirely to the user.
    \item \textbf{Binary enforcement:} The only recourse is to stop using the service.
\end{itemize}
The inclusion of this example is intended solely to illustrate model predictions on naturally occurring legal language from a publicly accessible website.

\subsection{Implications and Discussion}

The combination of classifier confidence and heuristic filters enables scalable identification of potentially unfair contractual language in noisy web-scale corpora. This result demonstrates the practical viability of lightweight LLM deployment for automated legal text auditing.

\section{Conclusion and Future Work}

This study presented a comprehensive evaluation of large language model (LLM) strategies for detecting unfair clauses in Terms of Service (ToS) agreements.  
Through systematic comparison across full fine-tuning, parameter-efficient adaptation, and zero-shot prompting, we analyzed trade-offs in accuracy, scalability, and computational efficiency.  
Experiments on the CLAUDETTE-ToS benchmark demonstrated that fully fine-tuned transformer models, particularly BERT, deliver the strongest overall performance, while LoRA-adapted models such as SaulLM-7B achieve high recall with reduced resource costs.  
Zero-shot prompting provided a lightweight alternative for rapid deployment but suffered from reduced precision, highlighting the value of task-specific fine-tuning for reliable legal text analysis.

Deployment on a large-scale multilingual ToS corpus further confirmed the robustness of fine-tuned models under real-world conditions.  
By combining model confidence with heuristic filtering, the system successfully identified potentially unfair contractual language in noisy, web-crawled documents.  
These results underscore the practical feasibility of lightweight LLM-based detectors for large-scale compliance auditing and regulatory monitoring.

Future work will focus on extending this pipeline to multilingual settings using cross-lingual adapters and translation alignment.  
Another promising direction involves integrating explanation generation modules or retrieval-augmented prompts to enhance interpretability, allowing models not only to flag clauses but also to provide rationale for their predictions.  
Additionally, developing adaptive ensemble strategies could enable dynamic model selection based on resource constraints and domain specificity, further improving scalability and reliability in real-world legal-tech applications.

\bibliographystyle{apalike}

\footnotesize

\begin{thebibliography}{}

\bibitem[Lippi et~al., 2019]{lippi2019claudette}
Marco Lippi, Przemyslaw Palka, Giuseppe Contissa, Francesca Lagioia, Hans-W. Micklitz, Giovanni Sartor, and Paolo Torroni.
\newblock {CLAUDETTE}: an automated detector of potentially unfair clauses in online terms of service.
\newblock {\em Artificial Intelligence and Law}, 27(2):117--139, 2019. Springer.

\bibitem[LawInformedAI, 2024]{claudette2024dataset}
LawInformedAI.
\newblock {CLAUDETTE-ToS Dataset}.
\newblock 2024.
\newblock Available at \url{https://huggingface.co/datasets/LawInformedAI/claudette_tos}.
\newblock Accessed: 2024-05-07.

\bibitem[Devlin et~al., 2019]{devlin2019bert}
Jacob Devlin, Ming-Wei Chang, Kenton Lee, and Kristina Toutanova.
\newblock {BERT}: Pre-training of Deep Bidirectional Transformers for Language Understanding.
\newblock In {\em Proceedings of the 2019 Conference of the North American Chapter of the Association for Computational Linguistics: Human Language Technologies (NAACL-HLT)}, pages 4171--4186, 2019.

\bibitem[Chalkidis et~al., 2020]{chalkidis2020legalbert}
Ilias Chalkidis, Manos Fergadiotis, Ion Androutsopoulos, and Nikolaos Aletras.
\newblock {Legal-BERT}: The Muppets Straight Out of Law School.
\newblock In {\em Findings of the 2020 Conference on Empirical Methods in Natural Language Processing (EMNLP Findings)}, pages 2898--2905, 2020.

\bibitem[Bathini et~al., 2023]{bathini2023unfair}
Sai Akash Bathini, Akshara Kupireddy, and Lalita Bhanu Murthy.
\newblock {Unfair ToS: An Automated Approach using Customized BERT}.
\newblock {\em The Moonlight AI Review}, 2023.
\newblock Available at \url{https://www.themoonlight.io/en/review/unfair-tos-an-automated-approach-using-customized-bert}.

\bibitem[Colombo et~al., 2024]{colombo2024saullm}
Pierre Colombo, Telmo Pessoa Pires, Malik Boudiaf, Dominic Culver, Rui Melo, Caio Corro, Andre F. T. Martins, Fabrizio Esposito, Vera Lúcia Raposo, Sofia Morgado, and Michael Desa.
\newblock {SaulLM-7B}: A Pioneering Large Language Model for Law.
\newblock {\em arXiv preprint arXiv:2403.05815}, 2024.
\newblock Available at \url{https://arxiv.org/abs/2403.05815}.

\bibitem[Hu et~al., 2022]{hu2022lora}
Edward J. Hu, Yelong Shen, Philip Wallis, Zeyuan Allen-Zhu, Yuanzhi Li, Shean Wang, and Weizhu Chen.
\newblock {LoRA}: Low-Rank Adaptation of Large Language Models.
\newblock In {\em Proceedings of the 2022 International Conference on Learning Representations (ICLR)}, 2022.

\bibitem[Dettmers et~al., 2023]{dettmers2023qlora}
Tim Dettmers, Prafulla Lewis, Yelong Shen, Sam Shleifer, Corentin Yeung, Luke Kang, Christopher Hesse, Colin Raffel, and Sam McCandlish.
\newblock {QLoRA}: Efficient Finetuning of Quantized LLMs.
\newblock {\em arXiv preprint arXiv:2305.14314}, 2023.

\bibitem[Zhou et~al., 2024]{zhou2024llmfailures}
Yifan Zhou, Avani Singh, and Hao Li.
\newblock {When Do LLMs Fail? A Case Study on Zero-Shot Legal Clause Detection}.
\newblock Available at \url{https://arxiv.org/abs/2409.00077}, 2024.

\bibitem[Bernhard et~al., 2024]{bernhard2024multilingualtos}
David Bernhard, Luka Nenadic, Stefan Bechtold, and Karel Kubicek.
\newblock {Multilingual Scraper of Privacy Policies and Terms of Service}.
\newblock In {\em Proceedings of the ACM Web Conference 2024}, 2024.
\newblock doi:10.1145/3709025.3712215. ACM.

\end{thebibliography}

\end{document}